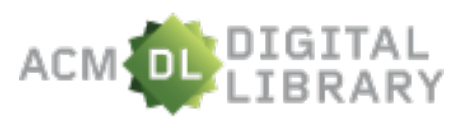
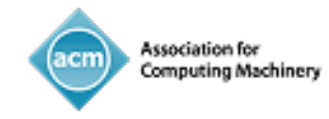
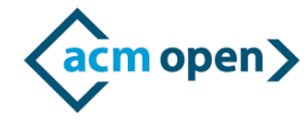

RESEARCH-ARTICLE

# MLFEF: Machine Learning Fusion Model with Empirical Formula to Explore the Momentum in Competitive Sports

**RUIXIN PENG**, Beijing Jiaotong University, Beijing, China

**ZIQING LI**, Huazhong University of Science and Technology, Wuhan, Hubei, China

# MLFEF: Machine Learning Fusion Model with Empirical Formula to Explore the Momentum in Competitive Sports

Ruixin Peng
Beijing Jiaotong University, Beijing, 100044, China
21723018@bjtu.edu.cn

Ziqing Li *
Huazhong University of Science and Technology, Wuhan, 430074, China
d202381502@hust.edu.cn

## ABSTRACT

Tennis is so popular that coaches and players are curious about factors other than skill, such as momentum. This article will try to define and quantify momentum, providing a basis for real-time analysis of tennis matches. Based on the tennis Grand Slam men's singles match data in recent years, we built two models, one is to build a model based on data-driven, and the other is to build a model based on empirical formulas. For the data-driven model, we first found a large amount of public data including public data on tennis matches in the past five years and personal information data of players. Then the data is preprocessed, and feature engineered, and a fusion model of SVM, Random Forrest algorithm and XGBoost was established. For the mechanism analysis model, important features were selected based on the suggestions of many tennis players and enthusiasts, the sliding window algorithm was used to calculate the weight, and different methods were used to visualize the momentum. For further analysis of the momentum fluctuation, it is based on the popular CUMSUM algorithm in the industry as well as the RUN Test, and the result shows the momentum is not random and the trend might be random. At last, the robustness of the fusion model is analyzed by Monte Carlo simulation.

## 1 INTRODUCTION

Tennis is a sport enjoyed by audiences around the world. In the 2023 Wimbledon men's singles final, the 20-year-old Spanish star Carlos Alcaraz defeated the 36-year-old Novak Djokovic. The game is full of suspense and twists, with the fluctuating dynamics of the entire game often attributed to "momentum," which refers to the psychological advantage a player or team gains over the course of a game, usually through consecutive wins. The concept of momentum in sports competition is often mentioned by athletes, coaches, and commentators alike, and therefore it is an important area of research. However, positive psychological dynamics appear to reflect psychological empowerment and concomitant changes in cognitive, affective, physiological parameters, and therefore performance. Therefore, how to define, measure and utilize momentum is a question worth exploring.

Psychological momentum is defined as "an additional or acquired psychological force that changes perceptions between people and affects an individual's mental and physical performance" [1]. Furthermore, the multidimensional momentum model [2] states that psychological momentum is caused by a series of events that trigger changes in cognitive, emotional, and physiological changes that influence athletes' behaviors and beliefs. These definitions suggest that psychological momentum may be a difficult concept to measure outside of the actual events that trigger these changes in mental states. Strategic motivation and psychological motivation, both of which have an impact on the outcome of best-of-three games [3]. The winner of all non-tie games performs worse than theoretically in the next game. This contradicts the expectation that psychological momentum will cause the winner of the previous set to perform better than theoretical probability. Instead, there is evidence of a "psychological reversal" in which winners underperform rather than outperform the theory [4]. Participants described several situations that they considered to be positive turning points. These situations are divided into two categories: directly related to the scoring system and not directly related. One situation that participants mentioned that was directly related to the scoring system was when they were about to lose the game, but they chose to take a risky shot and won the point [5].

It can be easily found that recent study on the players' psychological momentum is not deep, and most methods focused on the psychological methods with a few the-state-of-the-art technics. This paper has a very in-depth analysis of the psychological momentum of athletes, our contributions are as follows:

- Building a data-driven machine learning fusion model and an empirical formula model analyzed from the perspective of professionals.

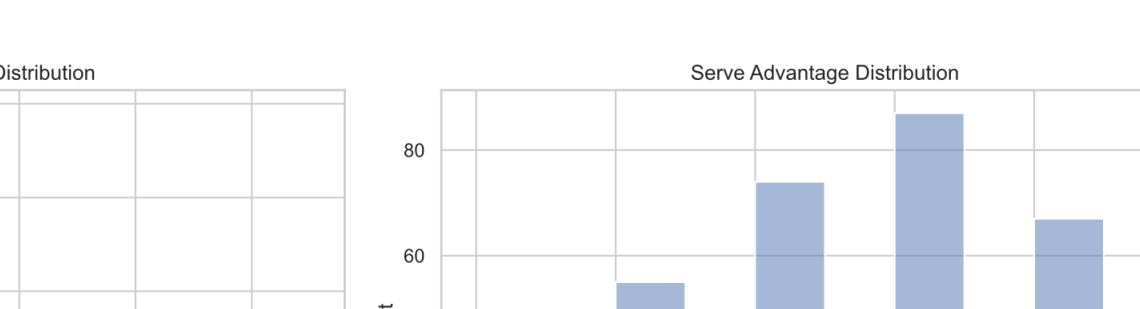
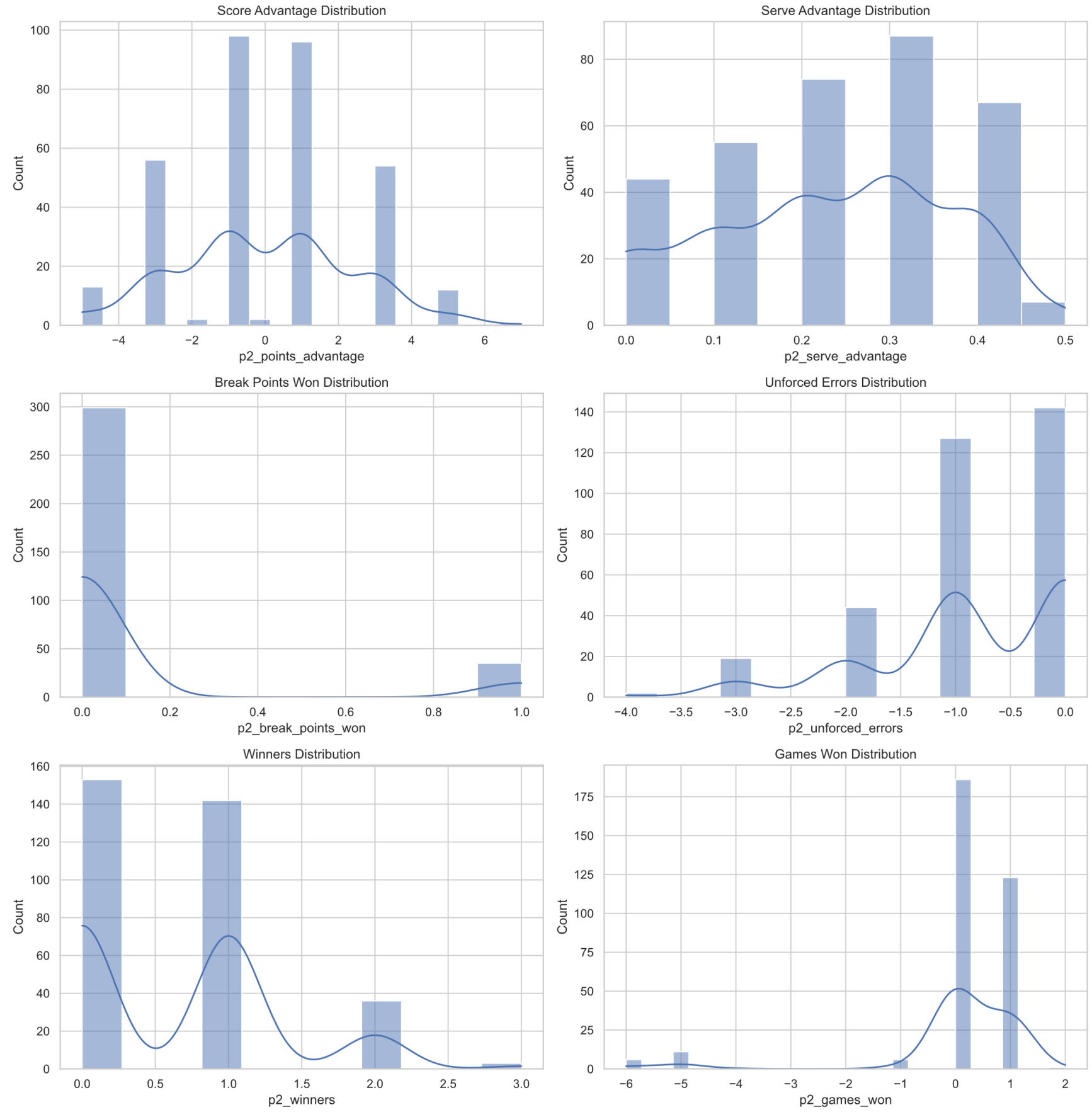

**Figure 1: Part of player's matching factors distribution.**

- Analyzing the feature importance and listing the importance ranking of all features, it is easy to find which has great influence on the match and the players.
- Visualization the changes of momentum throughout the match, and their turning points to indicate the swing in the competition.

## 2 MOMENTUM FACTOR ANALYSIS AND VISUALIZATION

In this part, there are some steps to analyze the statistics. We would have data preprocessing and feature engineering, data-driven model building, feature importance analysis and comparison and momentum visualization.

### 2.1 Data preprocessing and feature engineering

**Data Cleaning.** This step is missing Value Handling. First, we identify the missing values in the data, which is shown in Figure 1. Through this process, we observe that the column "return_depth" has 31 missing values. Further analysis of other matches, such as Carlos Alcaraz versus Nicolas Jarry, also reveals 48 missing values. Given that this type of missing value accounts for approximately 12% of the feature, it is not deemed necessary to impute them. Therefore, they are directly dropped.

**Data Imputation.** For other missing values, we perform imputation using the mode since these variables contain non-numeric values. While an alternative approach could involve imputing with interpolated values after one-hot encoding, it was not utilized in this instance. This decision is based on the observation that these features did not reveal significant importance in previous analyses and the potential complexity involved in the process.

**Data Transformation.** For certain non-numeric features, we performed **one-hot encoding**, converting categorical data into a format that is manageable for the model.

**PCA dimensionality reduction.** Principal Component Analysis (PCA) [6] is a mainstream data dimensionality reduction algorithm for addressing multicollinearity, aiming to retain maximum data information. In the process of linear transformation, the standard orthogonal basis serves as the eigenvector, and its eigenvalues are sorted. The first several eigenvectors correspond to the largest variance of data variables, containing the most information, known as principal components. Typically, an accumulated information retention rate of over 85% for principal components is considered satisfactory.

**Table 1: Features dimension reduction processing table.**

| Features to be reduced dimension | New features after dimensionality reduction |
| --- | --- |
| p1_points_won | points_won_meta |
| p2_points_won | |
| set_no | match_no_meta |
| game_no | |
| point_no | |
| p1_sets | sets_meta |
| p2_sets | |

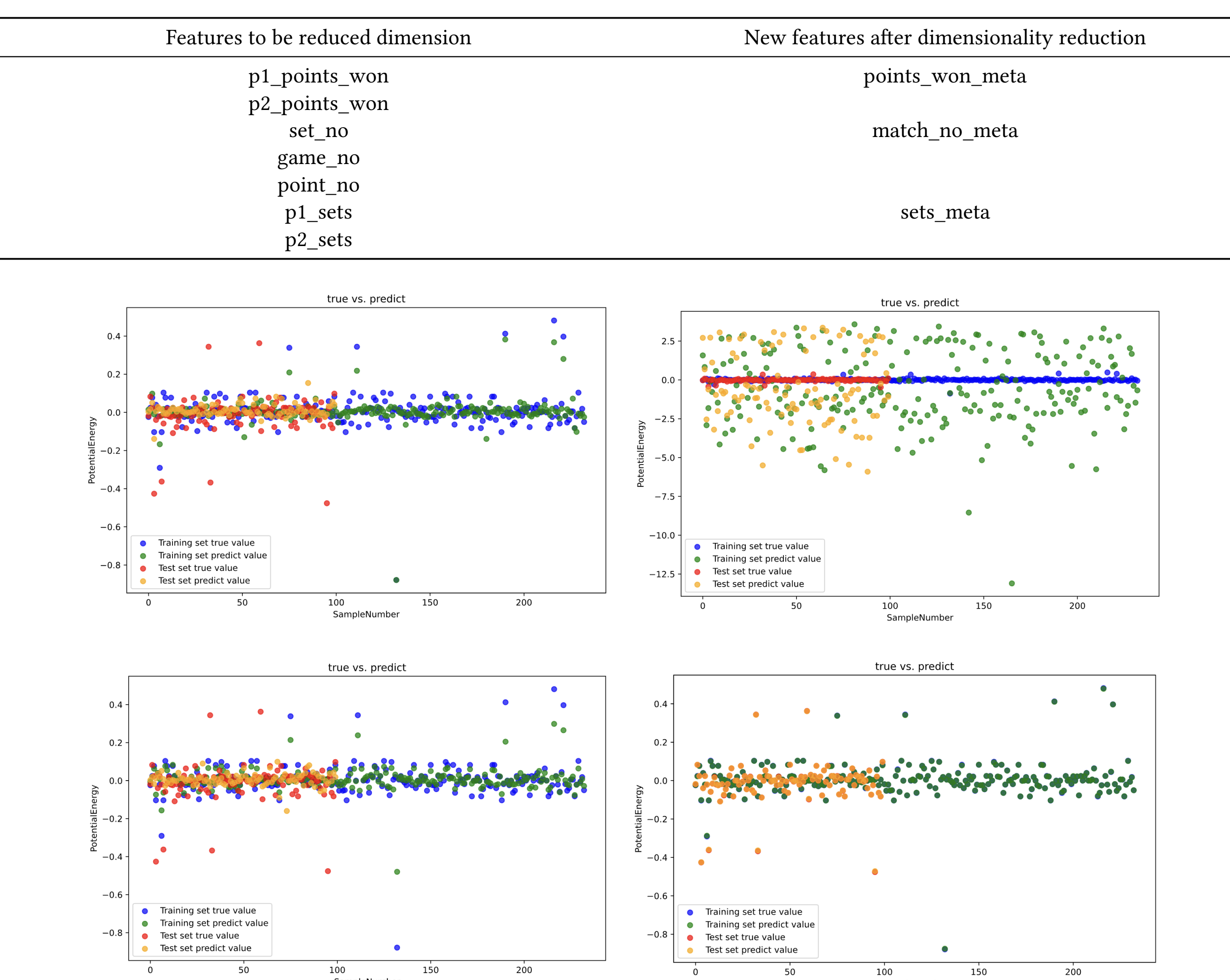

**Figure 2: The comparison by Neural Network, Random Forest, Adaboost and XGBoost.**

After multicollinearity diagnosis and dimensionality reduction, the number of features in the dataset has been reduced from 50 to 44. The fused data after PCA dimensionality reduction is shown in Table above. From Table 1, it can be observed that the fused data values themselves do not have a physical meaning. They are simply a compressed representation of the data before fusion, maximizing the reduction in information loss during compression. This is particularly suitable for machine learning models like XGBoost, which can uncover the underlying patterns in the data. The curve graph of the information retention rate after PCA dimensionality reduction for the fused features is shown in Figure 2.

## 2.2 Machine learning models make predictions

Firstly, it is important to note that this question involves a regression problem. Currently, mature machine learning algorithms for regression tasks include neural networks, random forest regression, Adaboost algorithm, XGBoost algorithm, etc. As previously introduced, for a single match, we will assess the predictive accuracy of these models. Regarding dataset partitioning, we randomly selected 70% as the training set and the remaining 30% as the test set.

In order to have a more accurate understanding of the data in the four graphs, the following table provides a more detailed description of their indicators.

**Table 2: Four different ML methods prediction results.**

| Model name | Training set mean absolute percentage error | Test set mean absolute percentage error | Mean absolute error | $r^2$ |
|---|---|---|---|---|
| Neural Network | 3.704 | 1.371 | 0.194340703 | -5.366658641 |
| Random Forest | 1.565 | 8.5 | 0.060978279 | -0.0892924 |
| Adaboost | 32.165 | 27.526 | 0.061488286 | -0.031616991 |
| XGBoost | 0.191 | 0.168 | 0.000876014 | 0.999851105 |

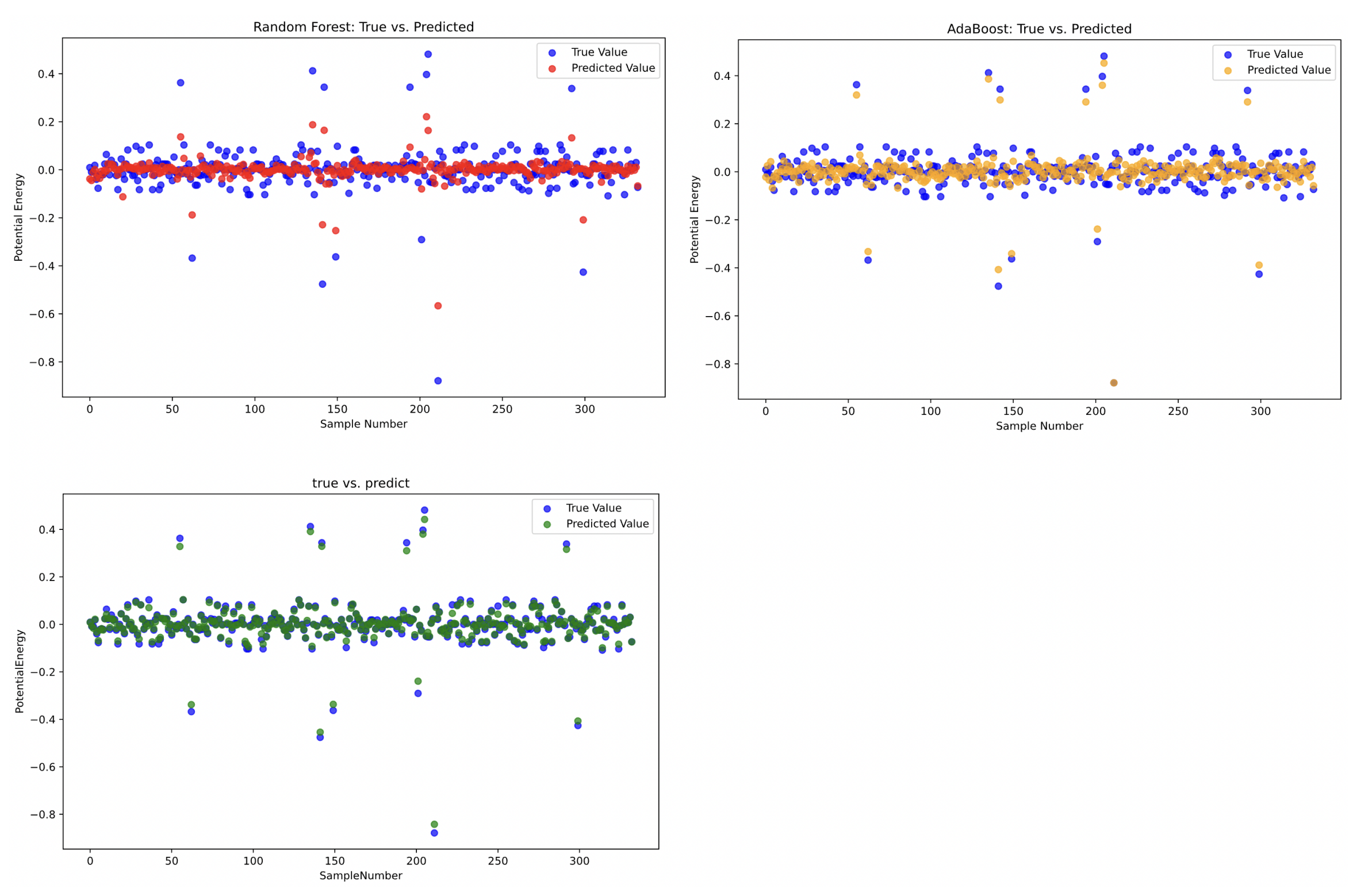

**Figure 3: XGBoost, Adaboost and Random Forests prediction results.**

Due to the condition should take more considerations on the mean absolute percentage error (MAPE) rather than r2 since MAPE has a more accurate description on the ML models. It can be observed from Table 2 that XGBoost has the best predictive performance, while Adaboost has the poorest predictive performance. XGBoost is an optimized distributed gradient boosting library designed to be efficient, flexible, and portable. XGBoost provides a gradient boosting framework that improves model performance and speed by utilizing advanced regularization techniques to prevent overfitting. Therefore, we ultimately plan to use **XGBoost** for prediction. The results are shown in the Figure 3 below:

The comparison results between predicted values and actual values are very good, validating the excellent predictive performance of XGBoost.

## 2.3 Feature importance analysis and comparison

Feature importance analysis is a crucial step in machine learning, aiming to assess the contribution of input features to the predictive performance of the model. This analysis not only helps understand the influencing factors behind the data but also guides feature selection to optimize the model's performance and interpretability. Common methods for feature importance analysis include model-based feature importance, permutation-based feature importance, SHAP values (SHapley Additive exPlanations), mutual information, and correlation coefficients (Pearson or Spearman).

By drawing the hot map of the feature in Figure 4, some important information can be used for us.

In competitive sports such as tennis, there may be strong correlations in the data of athletes from both sides. The Spearman correlation heatmap presented in our first question reflects this observation. To enhance the accuracy of our analysis, we opted to utilize SHAP (SHapley Additive exPlanations) for feature engineering. Grounded in game theory and local interpretation, SHAP is applicable across various models. Given its robust interpretability and flexibility, SHAP values are increasingly applied in the industry, particularly in scenarios where an in-depth understanding of the reasons behind model predictions is required. The results are as follows.

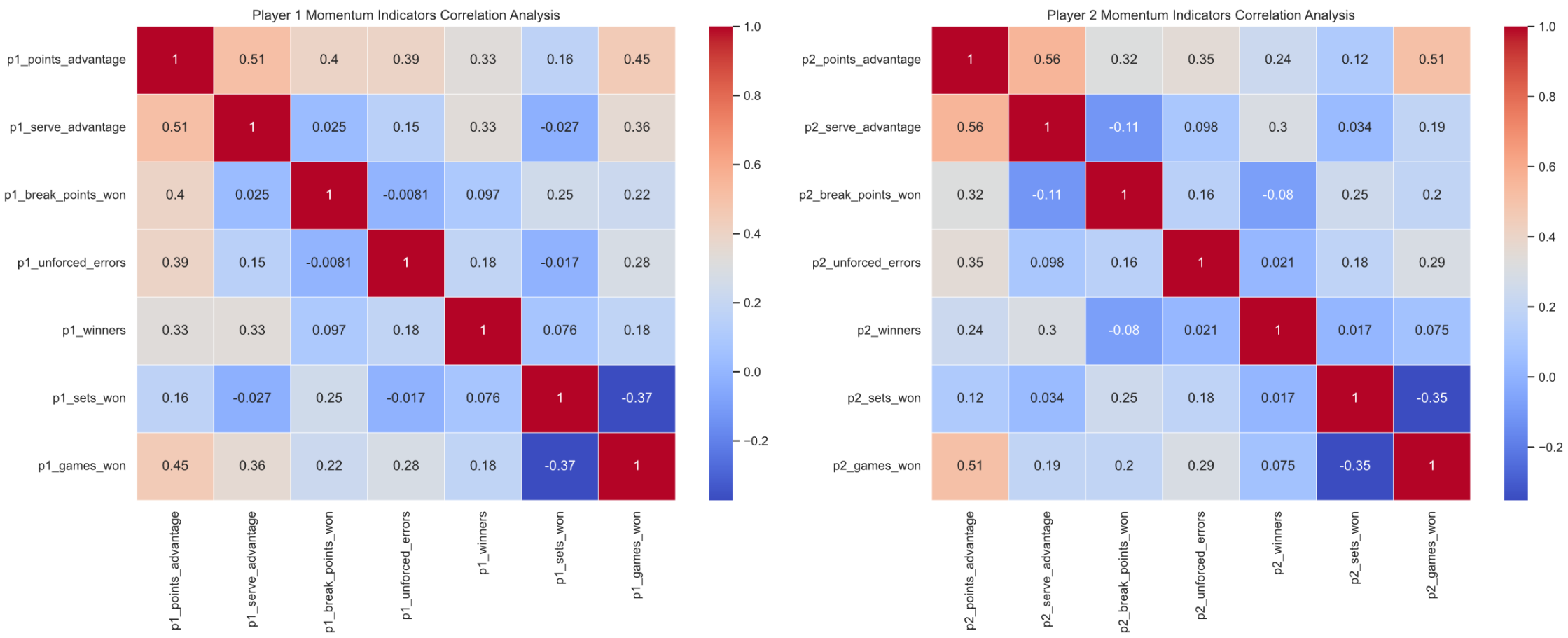

**Figure 4: Feature correlation hot map of both two players.**

The feature importance rankings by different methods are different as shown in Figure 5. The most significant values are **p2_distance** in ***XGBoost***, **speed_mph** in ***Adaboost*** and **rally_count** in ***Random Forests*** respectively. It can be observed that for Carlos Alcaraz, the running distance of Novak Djokovic is the most important feature, as it reflects the exhaustion of Djokovic's stamina. In terms of Alcaraz's own running distance, it is also a negatively correlated variable with his stamina. After excluding variables unrelated to the player, we find that speed_mph, rally_count, and others also exhibit significant correlations with important features. SHAP can have a clear visualization about the feature importance of each item in data, which can be shown in Figure 6 below.

## 2.4 Fusion Model

The machine learning algorithms used in the basic models of the fusion model in this article are support vector machine (SVM) suitable for classification problems, random forest (RF), a representative algorithm of gradient boosting algorithm, and extreme gradient boosting algorithm (XGBoost), a high-performance algorithm proposed in recent years. The whole diagram of the fusion model can be seen in Figure 7.

**Bayesian optimization of hyperparameters.** Use point_victor in the data set as the model label value and use the 35 retained parameters after data preprocessing and feature engineering as the feature parameters of the input model to establish a two-class prediction model [7]. Regarding the division of the data set, considering that the question requires that the model can be applied to one or more games, the finals of the event are used as the test set, the two semi-finals are used as the verification set, and the remaining data sets are used as the training set, so that the model can continuously predict the score for the entire game. By analysis, the optimal value of num_round is obtained at 126.

**Weighting Average.** The weighted average in the machine learning algorithm fusion method is to perform a linear weighted average of the output results of the basic model to generate a new fusion output result. The general form is as follows:

As shown in Figure 8, when each basic model uses default hyperparameters, the 10-fold cross-validation results show large differences. The performance of the RF model and the XGBoost model is close, but the XGBoost model performs better, while the SVM model per-forms relatively poorly. Calculating the mean of the 10-fold cross-validation results, the prediction accuracy of the SVM, RF, and XGBoost models on the basic model training set are 92.6%, 96.4%, and 97.5%, respectively, so their corresponding weight coefficients are 0.323, 0.336, and 0.340 respectively.

**Stacking.** Stacking refers to using the output results of multiple basic models as new input features, and then predicting them through a meta-model, thereby achieving hierarchical combination between models. [8] This article uses the logistic regression model as the meta-model. The final stacking function is shown as below:

$$
\begin{aligned}
F(x, y) = \ & 0.926 * SVM(x,\ y) + 0.964 * RF(x,\ y) \\
& + 0.975 * XGBoost(x,\ y)
\end{aligned} \tag{1}
$$

0.926, 0.964 and 0.975 are the weights of SVM, Random Forests and XGBoost merged into one large model respectively. By this stacking method, the model has a strong ability to predict the momentum in an excellent accuracy.

# 3 MOMENTUM VISUALIZATION AND FURTHER EXPLORATION

## 3.1 Momentum visualization

With the computed weights mentioned above, combined with the empirical formulas proposed earlier [9], we can derive the weights for each moment of every player. We plot the entire dataset's energy landscape to provide readers with a clear understanding of the changes in player energy throughout a match. The results are depicted in the following Figures 9

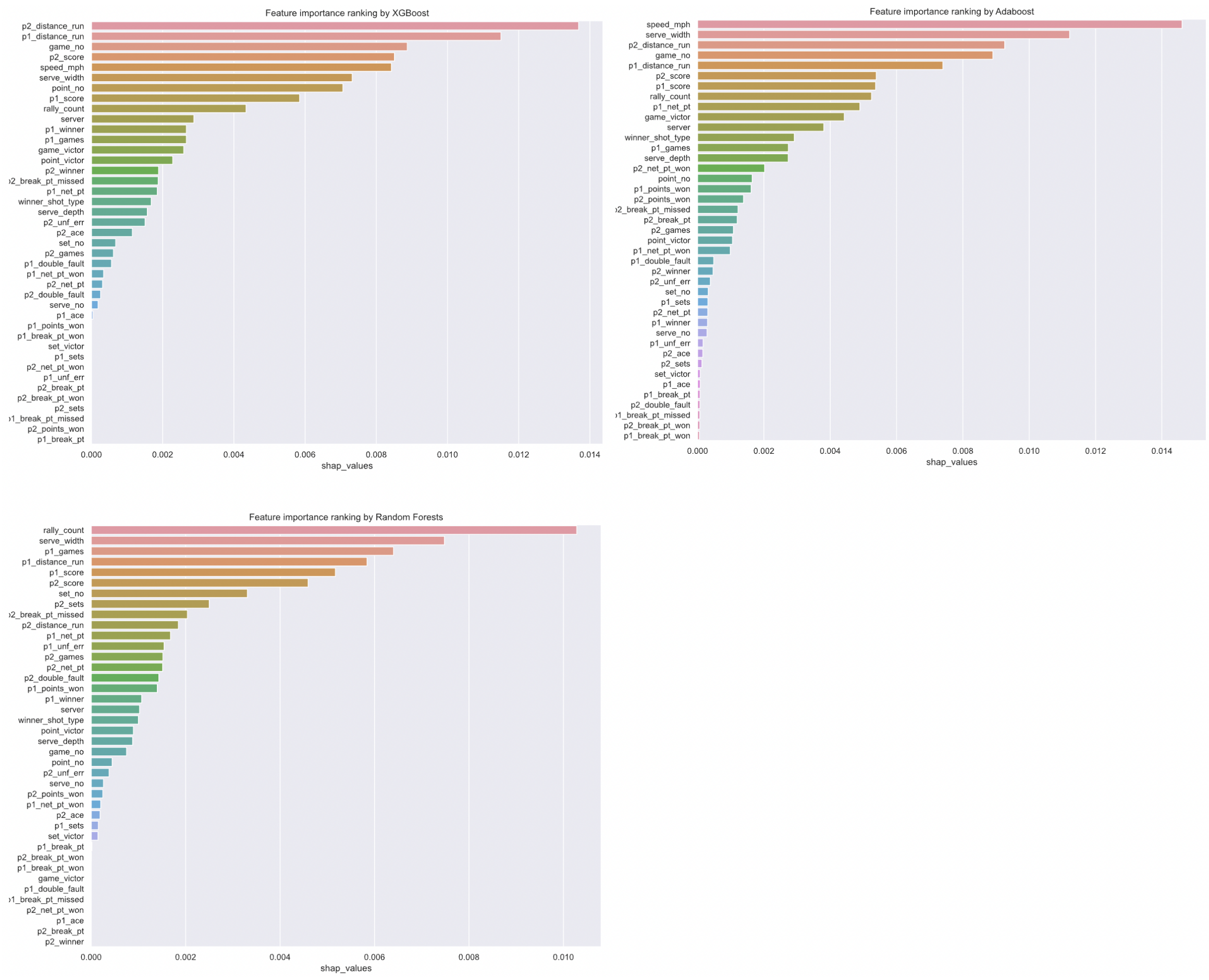

**Figure 5: Feature importance ranking by XGBoost, Adaboost and Random Forests.**

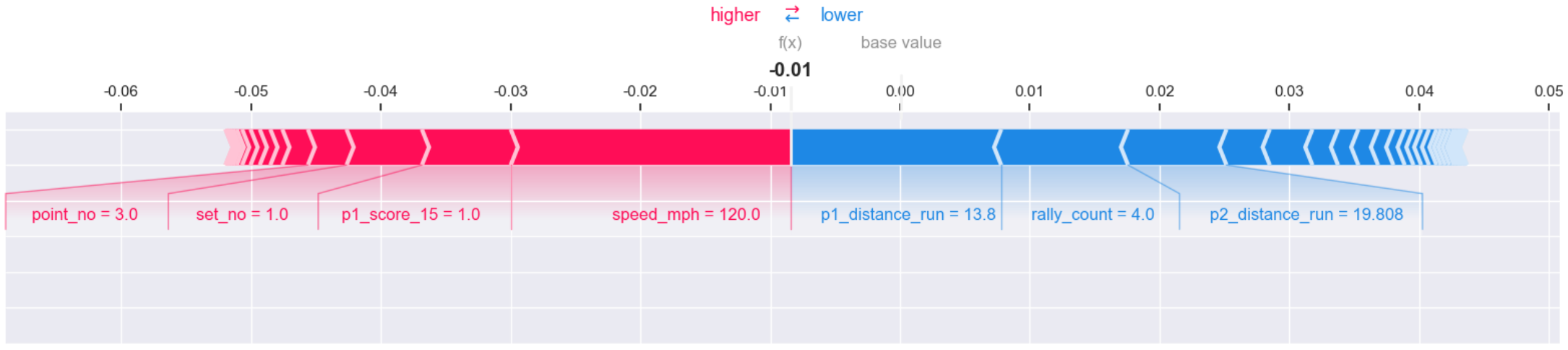

**Figure 6: An example about the feature importance in some time of the match.**

The stack area chart Figure 10 can visually show the cumulative change in momentum of two players over time, making it easy to see which player has the advantage at which time. In general, Carlos Alcaraz's potential energy performance is better than Novak Djokovic's most of the time.

## 3.2 The fluctuation of the game

This paper employs the Cumulative Sum (CUSUM), an excellent statistical method. CUSUM utilizes current and recent process data to detect slight changes or variability in the process mean. Represented as the "cumulative sum," CUSUM assigns equal weights to current and recent data. The design concept of CUSUM involves accumulating information from sample data and amplifying small sample deviations, thereby enhancing sensitivity in detecting minor shifts during the process. The CUSUM algorithm, known for its convenience, simple criteria, and ease of operation, finds widespread application in various fields, including industrial quality control, economics, automatic fault detection, finance, among others.

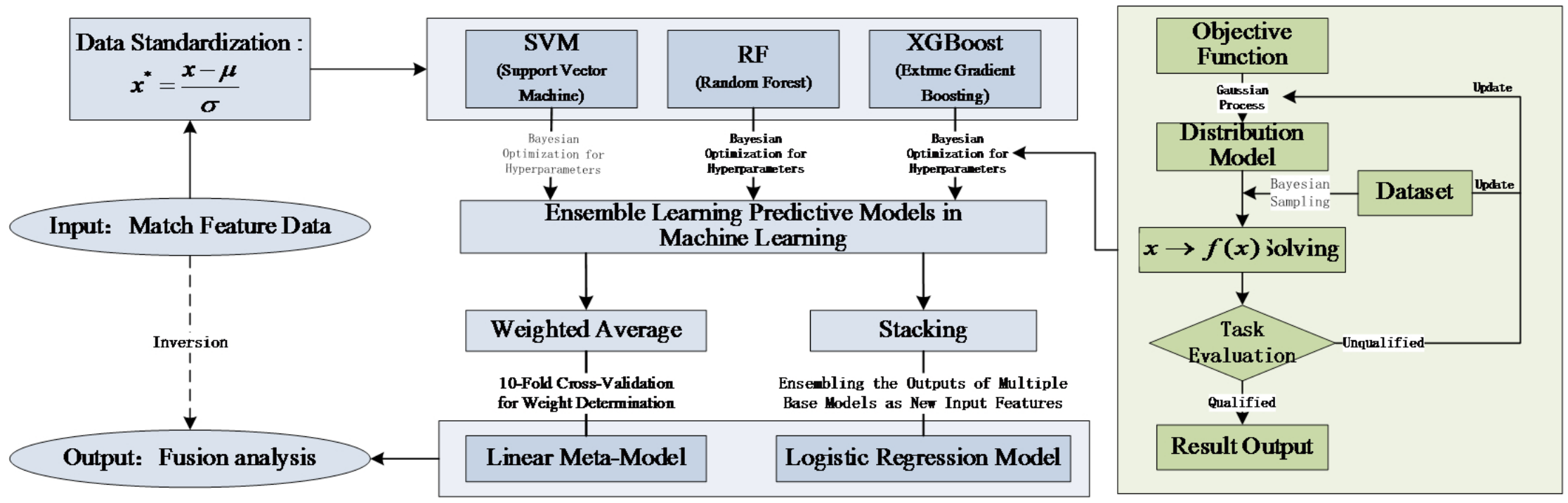

**Figure 7: Flowchart of algorithm fusion model construction.**

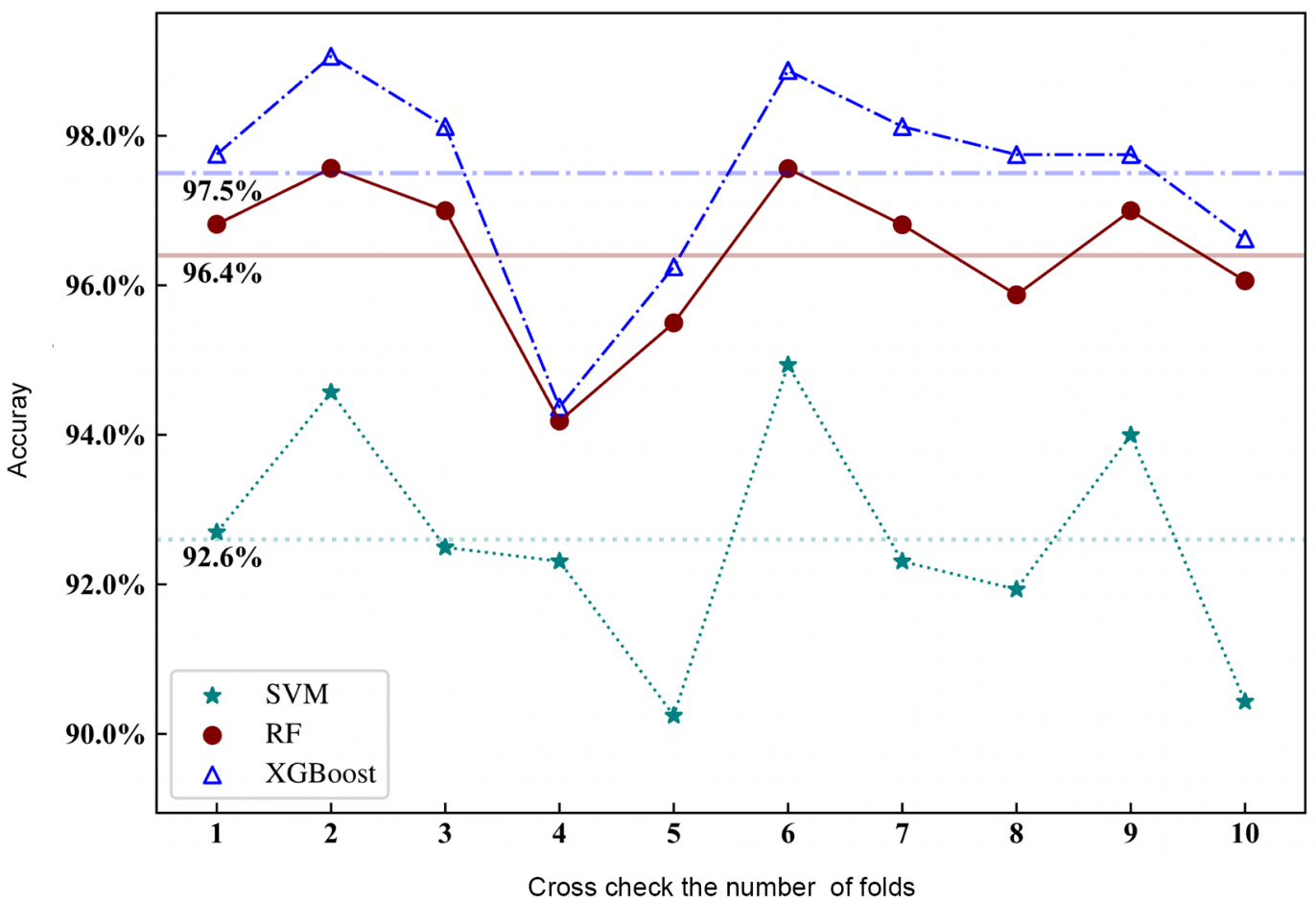

**Figure 8: 10-fold cross-validation results.**

This paper, by analyzing the momentum column defined in question 1 and utilizing the CUSUM algorithm, identifies turning points in the match, specifically swings in play and runs of success. In the definition of the CUSUM algorithm, a turning point is considered when the cumulative sum crosses the zero point again. Through the analysis of the provided data, all turning points in the match are marked in the Figure 11.

Using the CUSUM algorithm, this paper has identified all the turning points in the given match data. These turning points are defined as swings in play and runs of success, facilitating subsequent run tests to determine whether swings in play and runs of success by one player, as well as momentum, have an impact on the match outcome.

## 3.3 Run test on momentum and the fluctuation

By employing Python, two matches were randomly selected from the dataset, namely the 2023 Wimbledon matches numbered 1312 and 1601. For the selected match, 2023-wimbledon-1312, the momentum and turning point values for each player were computed, followed by a run test conducted on these four variables.

### 3.3.1 *Analysis of one game.*

According to the results shown in the table, the run test indicates a significant P-value of 0.000*** for variables p1_momentum and p2_momentum, demonstrating statistical significance. As a result, the null hypothesis is rejected, confirming that the data is non-random. On the other hand, for variables p1_turning_points and p2_turning_points, the P-values are 0.943 and 0.696, respectively, which are not statistically significant, leading to the failure to reject the null hypothesis. Consequently, the data for variables p1_turning_points and p2_turning_points is considered random. For the selected match, 2023-wimbledon-1601, the momentum and turning point values for each player were computed, and run tests were conducted on these four variables. The results of the analysis are presented in the Table 3 below.

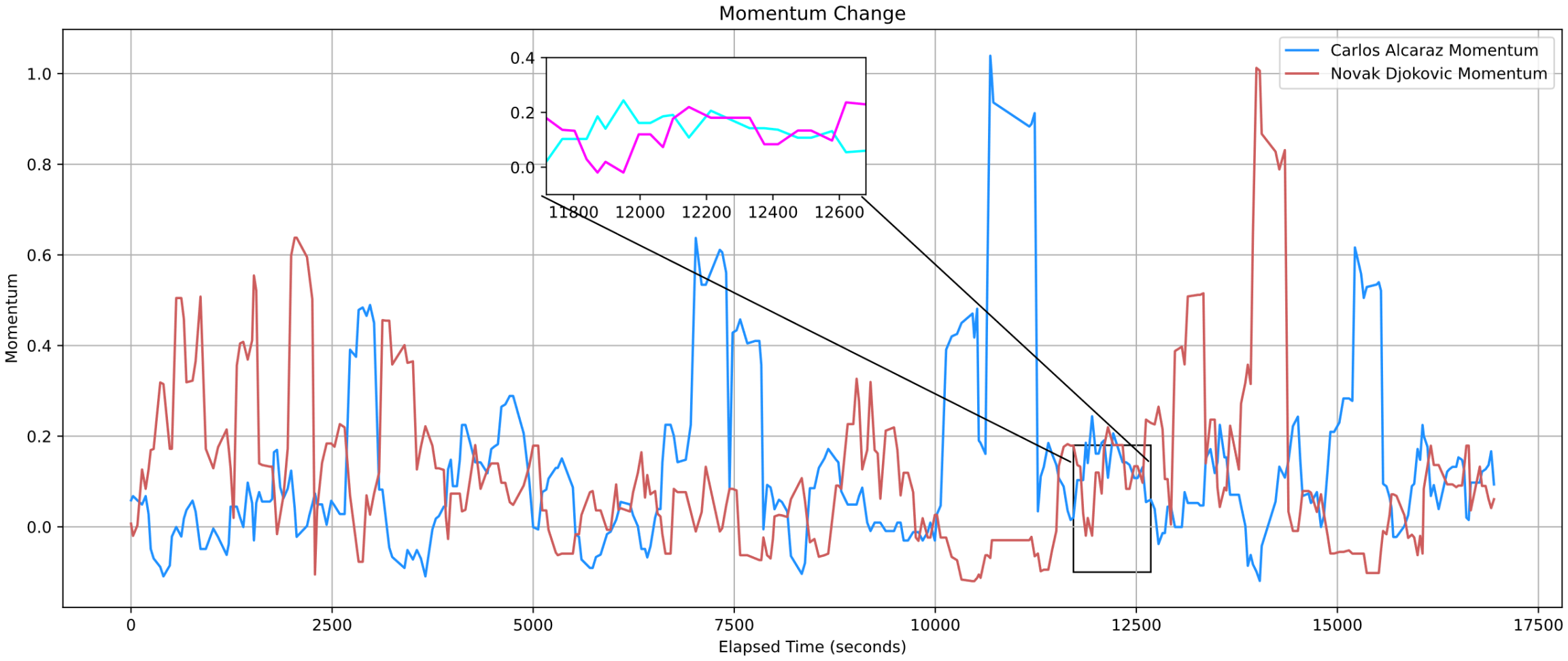

**Figure 9: Momentum change.**

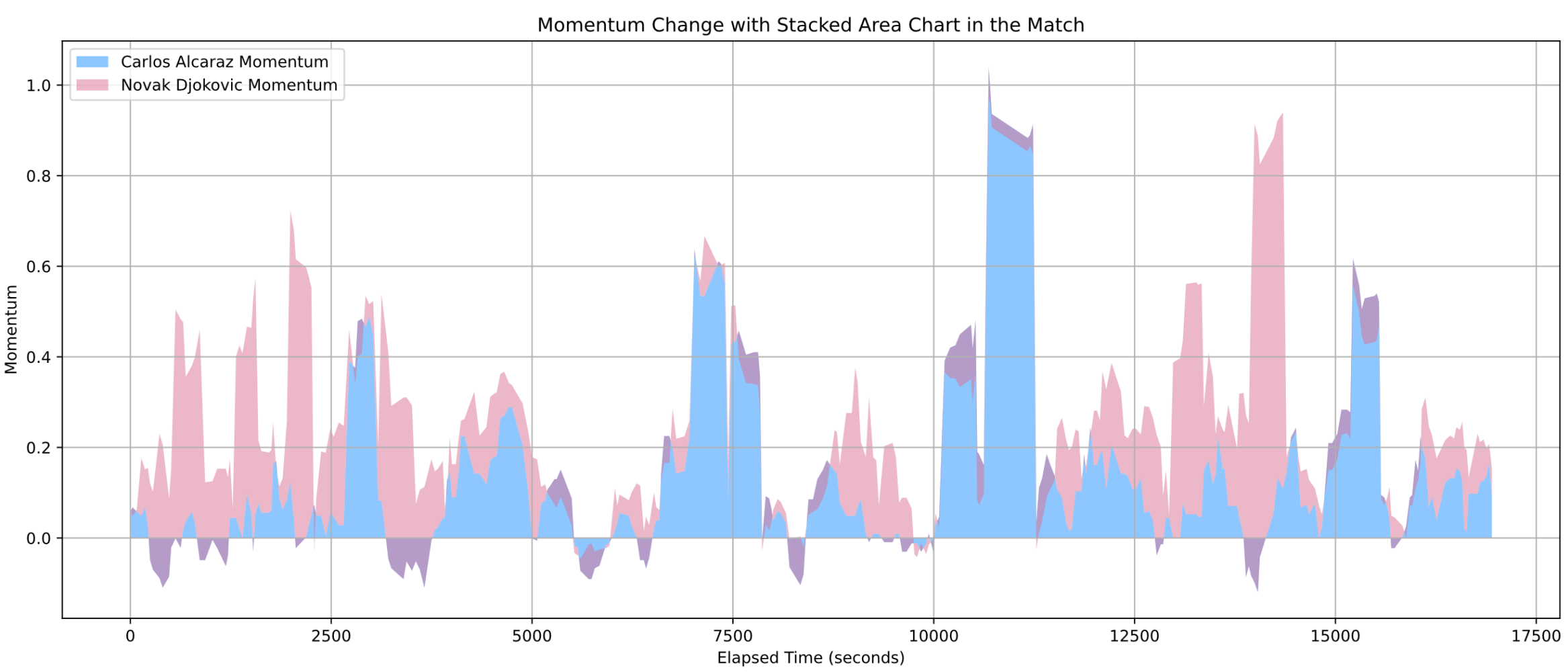

**Figure 10: Momentum changes with stacked area chart.**

**Table 3: Run test results of one game.**

| colume_name | sample_size | z | P-value |
|---|---|---|---|
| p1_momentum | 170 | -9.385 | 0.000*** |
| p2_momentum | 170 | -8.923 | 0.000*** |
| p1_turning_points | 170 | -0.071 | 0.943 |
| p2_turning_points | 170 | -0.391 | 0.696 |

From this Table 3, the only conclusion is the momentum and turning points might have the different behaviors. It can not be directly deduced that they have distinct features during one game and the accurate conclusion should be drawn after the analysis of the whole multiple games next.

*3.3.2 Analysis of multiple games.* We expanded our analysis to cover the entire official dataset, conducting a comprehensive examination of the data for the entire 2023 Wimbledon tournament. We established a function that outputs 1 if the run test's p-value for a match (e.g., 2023-wimbledon-1601) is less than 0.05, and 0 otherwise. The run test results for the 31 matches in the dataset are presented in the descriptive statistics table 4.

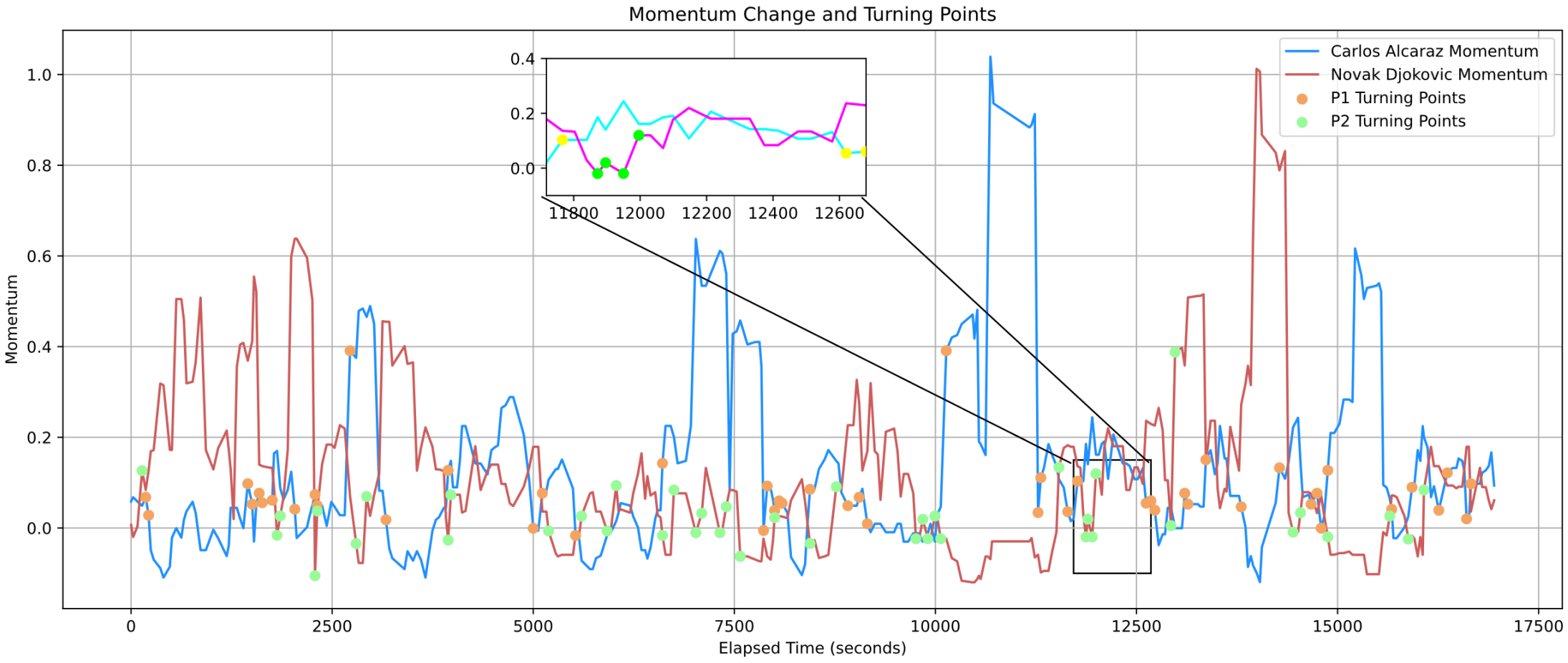

**Figure 11: Potential energy and turning points visualization.**

**Table 4: Run test results of multiple games.**

| | p1_momentum | p2_momentum | p2_turning_points | p2_turning_points |
|---|---|---|---|---|
| count | 31.0 | 31.0 | 31.000000 | 31.000000 |
| mean | 1.0 | 1.0 | 0.387097 | 0.258065 |
| std | 0.0 | 0.0 | 0.495138 | 0.44803 |
| min | 1.0 | 1.0 | 0.000000 | 0.000000 |
| max | 1.0 | 1.0 | 1.000000 | 1.000000 |

Analyzing this Table 4, the study concludes that the momentum for both Player 1 and Player 2 is not random. Furthermore, 38.7% of the turning points for Player 1 and 25.8% for Player 2 are considered non-random. Therefore, the study asserts that, in the context of the matches analyzed, players' momentum is not random, while turning points are random to a certain extent.

## 4 MODEL ROBUSTNESS ANALYSIS

Monte Carlo simulation is essentially a method that uses random numbers to solve computational problems and is often used to test the robustness of the model under certain parameter perturbations [10]. For the momentum prediction model (taking P1Momentum as an example), the training set and the test set are randomly divided into the original data set at a ratio of 7:3, and 1000 Monte Carlo simulations are generated using the random numbers of the divided data set, allowing each model to continuously learn random Divide the training set and predict the corresponding test set. The probability density distribution diagram of the evaluation index of the Monte Carlo simulation results is shown in the figure 12.

As can be seen from Figure 12, the overall performance of each model is good. The distribution of the RF model is closest to 1, and the distribution peaks of the three models are around 0.997, 0.987, and 0.97 respectively. The RF model has the smallest overall mean distribution, followed by the XGBoost and SVM models. In the distribution, the peak values of the XGB model and the RF model are close to each other, but the peak value of the RF model is higher, and the performance of the SVM model is poor. The RF and XGBoost models have strong overall performance and good robustness, while the overall performance of the SVM model is relatively poor. The fusion model proposed in this article is based on these three basic models and is theoretically more robust.

## 5 CONCLUSION

This paper mainly explores the momentum of athletes in tennis and has done a lot of detailed work. It uses a high-quality input data. In terms of data set preprocessing and feature engineering, we have carried out a lot of detailed and logical work, including data cleaning, multicollinearity diagnosis, PCA dimensionality reduction, significance analysis based on binary logistic regression, feature importance analysis based on decision tree model, and data standardization. The fusion model shows a great prediction result. Then the momentum visualization is displayed in two different modes which can help people easily understand the process of momentum through the match. CUMSUM algorithm is applied to detect the fluctuation in the competition and RUN TEST is used to confirm our ideas, that the momentum is not random but the turning points in matches is hard to predict based on previous behavior.

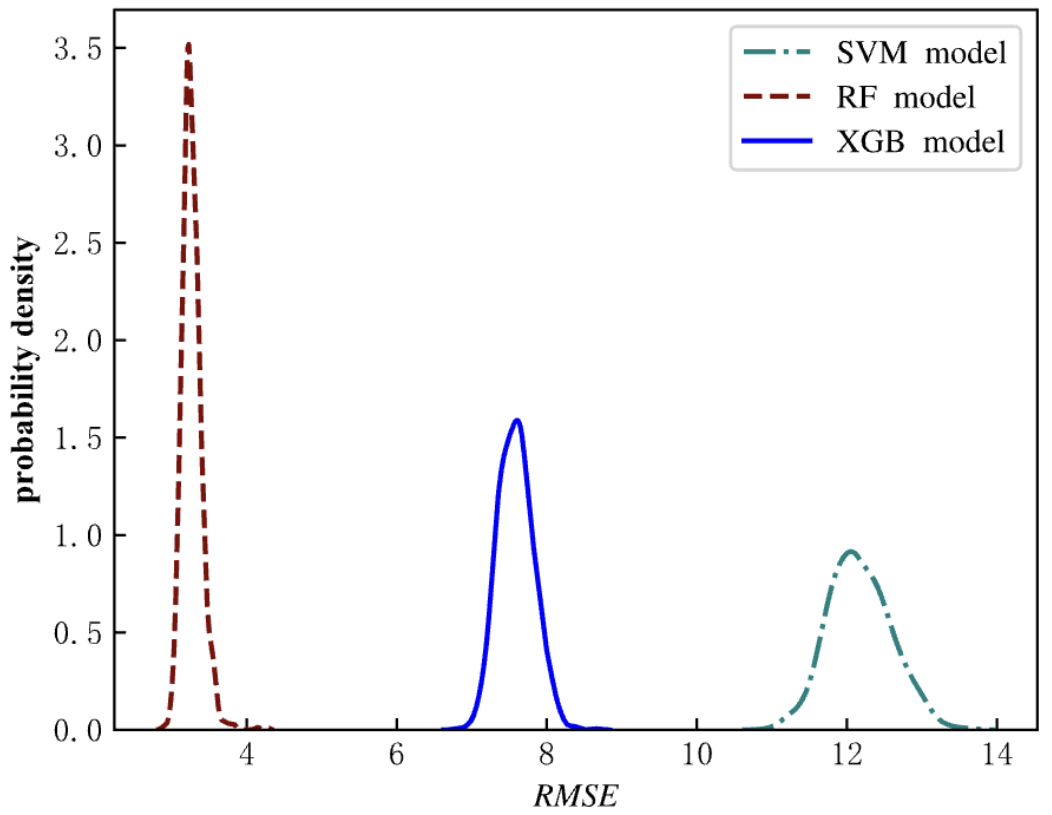

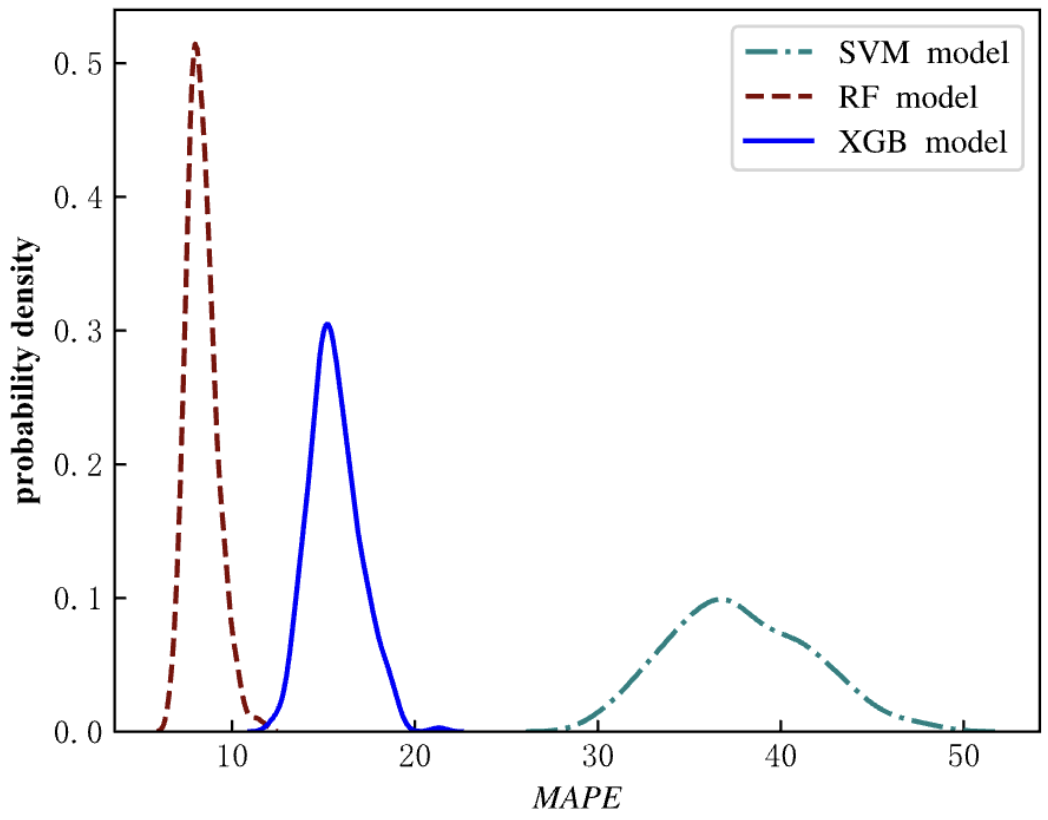

**Figure 12: Probability density of Monte Carlo simulation evaluation index.**

## ACKNOWLEDGMENTS

Acknowledgments to some great students' work. They are Hao Yin (BJTU), Xinrui Zhu (BJTU) and the first author Ruixin Peng. Their participation in COMAP-MCM competition provides a great idea for this paper. Their work is significant and excellent to analyze the tennis match momentum. Last but not the least, sincere thanks to Haodong Lin (CUMT), his rich modeling experience and paper writing skills offer us many important guidance to finish our work. For some future work, we would like to combine the analysis of the sports betting [10] to explore whether they have some conections.